\documentclass{article}

\usepackage[preprint, nonatbib]{neurips_2024}

\usepackage[utf8]{inputenc} 
\usepackage[T1]{fontenc}    
\usepackage{hyperref}       
\usepackage{url}            
\usepackage{booktabs}       
\usepackage{amsfonts}       
\usepackage{nicefrac}       
\usepackage{microtype}      
\usepackage{xcolor}         

\usepackage{graphicx}
\usepackage{makecell}
\usepackage{multirow} 

\usepackage{amsmath}
\usepackage{algorithm}
\usepackage{algorithmic}
\usepackage{adjustbox}

\title{FIMP: Foundation Model-Informed Message Passing \\ for Graph Neural Networks}

\author{%
  Syed A. Rizvi\thanks{Equal contribution.} \\
  Yale University\\
  \texttt{syed.rizvi@yale.edu} \\
  \And
  Nazreen Pallikkavaliyaveetil$^*$ \\
  Yale University\\
  \texttt{nazreen.pm@yale.edu} \\
  \And
  David Zhang \\
  Yale University\\
  \And
  Zhuoyang Lyu \\
  Brown University\\
  \And
  Nhi Nguyen \\
  New York University\\
  \And
  Haoran Lyu \\
  MIT\\
  \And
  Benjamin Christensen \\
  ETH Zurich\\
  \And
  Josue Ortega Caro \\
  Yale University\\
  \And
  Antonio H. O. Fonseca \\
  Yale University\\
  \And
  Emanuele Zappala \\
  Idaho State University\\
  \And
  Maryam Bagherian \\
  University of Massachusetts Boston\\
  \And
  Christopher Averill \\
  Baylor College of Medicine\\
  \And
  Chadi G. Abdallah \\
  Baylor College of Medicine\\
  \And
  Amin Karbasi \\
  Yale University\\
  \And
  Rex Ying \\
  Yale University\\
  \And
  Maria Brbic \\
  EPFL\\
  \AND
  Rahul M. Dhodapkar \\
  University of Southern California\\
  \texttt{rahul.dhodapkar@med.usc.edu} \\
  \And
  David van Dijk \\
  Yale University\\
  \texttt{david.vandijk@yale.edu} \\
}

\begin{document}

\maketitle

\begin{abstract}
  Foundation models have achieved remarkable success across many domains, relying on pretraining over vast amounts of data. Graph-structured data often lacks the same scale as unstructured data, making the development of graph foundation models challenging. In this work, we propose Foundation-Informed Message Passing (FIMP), a Graph Neural Network (GNN) message-passing framework that leverages pretrained non-textual foundation models in graph-based tasks. We show that the self-attention layers of foundation models can effectively be repurposed on graphs to perform cross-node attention-based message-passing. Our model is evaluated on a real-world image network dataset and two biological applications (single-cell RNA sequencing data and fMRI brain activity recordings) in both finetuned and zero-shot settings. FIMP outperforms strong baselines, demonstrating that it can effectively leverage state-of-the-art foundation models in graph tasks.
\end{abstract}

\section{Introduction}
\label{sec:intro}

Foundation models have emerged as a new paradigm in artificial intelligence, shifting from narrow, task-specific training to large-scale pretraining of more generalized models \cite{brown2020language}. Through pretraining on vast amounts of data, foundation models serve as a base model which can be adapted to a variety of downstream tasks \cite{bommasani2021opportunities}. Pretraining is typically done in self-supervised fashion through autoregressive language modeling \cite{radford2018improving} or masked language/image modeling \cite{devlin2018bert, chen2020uniter}. Standard foundation models have emerged in fields such as Natural Language Processing (NLP) with BERT \cite{devlin2018bert}, GPT-3 \cite{brown2020language}, and CLIP \cite{radford2021learning}, as well as in Computer Vision (CV) \cite{yuan2021florence}. After pretraining, these models exhibit strong positive transfer to downstream applications, outperforming task-specific models \cite{gururangan2020don}. More recently, fields such as single-cell RNA sequencing and neuroscience have also seen the emergence of large-scale foundation models such as scGPT \cite{cui2023scgpt}, Geneformer \cite{theodoris2023transfer}, and BrainLM \cite{ortega2023brainlm}.

In contrast to the success of foundation models in domains such as language and vision, replicating large-scale pretraining for foundation models on graphs has proved more challenging. Often, there is less graph-structured data publicly available relative to the amount of unstructured data, with cost or technological limits hindering our ability to scale graph-structured data to rival unstructured data. In single-cell RNA sequencing (scRNAseq) data, for instance, technological advancements have fueled an exponential rise in the amount of unstructured single-cell transcriptomes available for analysis \cite{svensson2018exponential}, however spatial sequencing methods which preserve spatial organization of the cells within the tissue are not as well-developed. Furthermore, tokenization differences separate traditional Graph Neural Networks (GNNs) from pretrained transformer-based models. Traditional GNNs represent nodes using a single node embedding vector, whereas transformers tokenize input samples into into sequences of feature tokens, which places token representation at the feature level rather than the node level. Notable examples include gene tokenization in single-cell foundation models such as scGPT \cite{cui2023scgpt} and Geneformer \cite{theodoris2023transfer}, as well as image patching used in Vision Transformers (ViTs) \cite{dosovitskiy2020image, he2022masked}. \textbf{Bridging the gap between traditional GNNs and pretrained foundation models, and by extension unstructured and structured data, remains an open challenge.}

\begin{figure}
  \centering
  \includegraphics[width=1.0\linewidth]{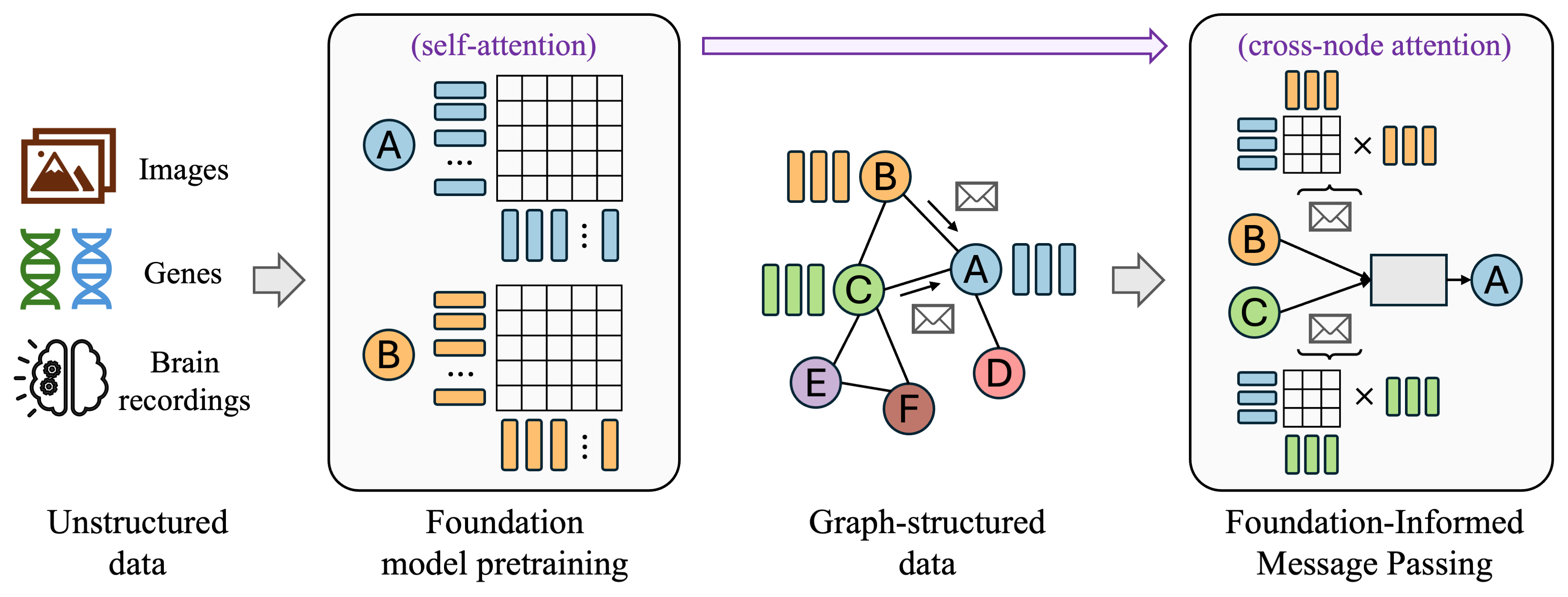}
  \caption{The proposed framework for FIMP. Pre-existing foundation models, pretrained on vast amounts of unstructured data, are repurposed into message creation modules by adapting their self-attention layers for cross-node attention between node feature sequences.}
  \label{fig:FIMP_operator_figure1}
\end{figure}

Existing works have begun leveraging pretrained foundation models in graph-based tasks, focusing on the application of LLMs on text-attributed graphs. One-for-All \cite{liu2023one} used LLMs as an encoding module for text-attributed graphs, and trained a GNN model to do node, edge, and graph-level classification. Talk Like a Graph \cite{fatemi2023talk}, NLGraph \cite{wang2023can}, and GPT4Graph \cite{guo2023gpt4graph} evaluated LLM reasoning capabilities on graph reasoning tasks, proposing benchmark datasets for simple graph tasks. These methods, however, focus only on improving performance on text-attributed graphs using LLMs, \textbf{leaving non-textual foundation models and domains underexplored in graph-based settings}.

To address these challenges, we propose Foundation-Informed Message Passing (FIMP), a message-passing framework for GNNs which repurposes the attention layers of pretrained non-textual foundation models for message-passing on graphs. We first unify node entity tokenization between foundation models and GNNs, and develop a cross-node attention-based message creation module which can be learned from scratch or initialized from pretrained foundation models. We evaluate our model on a real-world street-view image dataset \cite{antequera2020mapillary}, several spatial transcriptomics datasets, and fMRI brain activity recordings, incorporating state-of-the-art (SOTA) foundation models for images \cite{dosovitskiy2020image}, scRNAseq \cite{cui2023scgpt}, and brain recordings \cite{ortega2023brainlm} as message creators. FIMP demonstrates improvements over strong baselines, demonstrating that non-textual foundation models can be effectively leveraged in graph-based tasks. Additionally, FIMP demonstrates zero-shot embedding capabilities on image networks by leveraging pretrained ViTs \cite{dosovitskiy2020image}, matching GNNs trained from scratch. This demonstrates the potential of non-textual foundation models for zero-shot applications on graphs.

\textbf{Contributions.} In summary, the key contributions of our work are: 

\begin{enumerate}
    \item Introducing FIMP, a message passing framework which leverages pretrained non-textual foundation models for graph-based tasks.
    \item Evaluating FIMP across image, spatial transcriptomics, and brain activity recording datasets, incorporating SOTA foundation models in each domain as message creators.
    \item Demonstrating zero-shot embedding capabilities of FIMP with ViTs on image networks.
\end{enumerate}

\section{Preliminaries}
\label{sec:preliminaries}

\subsection{Graph Neural Networks}

Graph Neural Networks are a versatile class of neural network architectures which operate over graph-structured data. The core idea of GNNs is to learn node and/or edge attributes through iterative local aggregation steps, which is commonly implemented through Message-Passing Neural Networks (MPNNs) \cite{gilmer2017neural}. Below we define our notations for describing GNNs.

Let \(G = (V, E)\) denote a graph with a set of nodes $V$ and edges $E$. Each node has an input feature vector \(\vec{x}_i \in \mathbb{R}^f\), where \(f\) is the number of input features per node. GNNs iteratively pass messages between neighboring nodes, and in the process use both node features and graph structure to learn node representations \(\vec{h}_i \in \mathbb{R}^d\), where $d$ is the hidden dimension of node embeddings. After \(K\) message-passing iterations, node representation \(\vec{h}_i\) will contain information from its \(K\)-hop neighborhood within the graph. The general update rule for the \(k\)-th layer of a GNN can be represented as follows:

\begin{equation}\label{eq:gnn_neighbor_agg}
    \vec{h}_{\mathcal N(i)}^{(k)} = \texttt{AGGREGATE}^{(k)}\big(\big\{ \vec{h}_j^{(k-1)}, \forall j \in \mathcal N(i) \big\}\big)
\end{equation}

\begin{equation}\label{eq:gnn_node_update}
    \vec{h}_i^{(k)} = \texttt{COMBINE}^{(k)}\big( \vec{h}_i^{(k-1)}, \vec{h}_{\mathcal N(i)}^{(k)}),
\end{equation}

where \(\mathcal N(i)\) denotes the neighborhood of node \(i\) and \(h_i^{(k)}\) is the representation of node \(i\) in layer \(k\). The choice of \(\texttt{AGGREGATE}\) and \(\texttt{COMBINE}\) vary among different GNN architectures, with the constraint that \(\texttt{AGGREGATE}\) should be a permutation-invariant aggregator. A readout function is used to map learned node representations into predictions for feature, node, or graph-level tasks.

\subsection{Attention-based GNNs}

Graph Attention Networks (GATs) \cite{velivckovic2017graph} proposed to learn attention coefficients between neighboring nodes, replacing the \(\texttt{AGGREGATE}\) function in equation \ref{eq:gnn_neighbor_agg} with a linear combination of neighboring node vector embeddings weighted by normalized attention coefficients:

\begin{equation}\label{eq:gat_attention_equation_1}
    e_{ji} = a(\textbf{W} \vec{h}_i || \textbf{W} \vec{h}_j)
\end{equation}

\begin{equation}\label{eq:gat_attention_equation_2}
    \alpha_{ji} = \text{softmax}_j(e_{ji})
\end{equation}

where $\alpha_{ji}$ represents the final normalized attention coefficient between nodes $i$ and $j$, $a$ is a learned attention mechanism shared across all node pairs, and \textbf{W} represents a shared weight matrix.

We note that FIMP is fundamentally different from GATs and other attention-based GNNs such as graph transformers (covered in Section \ref{sec:related_works}). In GATs and graph transformers, nodes are represented by a single embedding vector \(\vec{h}_i\), and a scalar attention value is learned between neighboring nodes. In FIMP, nodes are tokenized into sequences of feature vectors \(H_i\), and cross-node attention between the feature sequences of neighboring nodes is used to create messages which are passed along edges. Further details of FIMP's novel message creation method are covered in Section \ref{sec:methodology}.

\subsection{Foundation models}

Foundation models are generalized Deep Learning models which have been pretrained on large amounts of data, and which can be finetuned for a variety of downstream tasks. In this work, we focus on non-textual foundation models, which define a tokenization procedure for continuous-valued data and typically do pretraining using a masked reconstruction objective. In single-cell RNA sequencing, for instance, scGPT \cite{cui2023scgpt} tokenizes an input cell as a sequence of gene tokens, and learns a gene embedding table analogous to word embeddings learned in LLMs. Pretraining is done through a masked gene expression prediction objective. In the image domain, ViT-based architectures \cite{dosovitskiy2020image, he2022masked} encode images as a sequence of patches, and similarly for fMRI brain activity recordings, BrainLM \cite{ortega2023brainlm} tokenizes segments of brain activity signal per brain region into tokens.


\section{Foundation-Informed Message Passing}
\label{sec:methodology}

We propose a novel message-passing framework, depicted in Figure \ref{fig:FIMP_operator_figure1}, that uses pretrained non-textual foundation models to generate messages between neighboring nodes in a graph. This leverages the pretrained knowledge of the foundation model to inform message-passing, allowing for pretraining on unstructured data before training on less-abundant graph-structured data.

\begin{figure}
  \centering
  \includegraphics[width=1.0\linewidth]{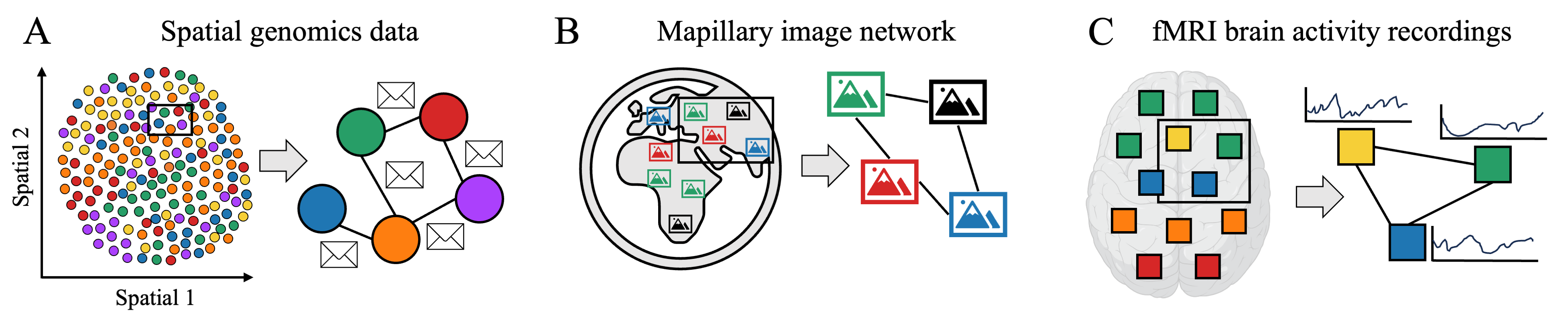}
  \caption{Graph structure present in real-world datasets. (A) In spatially resolved RNA transcriptomics, cells are connected to adjacent cells in the 2D tissue section. (B) In the Mapillary street-view image dataset \cite{antequera2020mapillary}, images form a geographical proximity graph based on latitude and longitude coordinates. (C) For fMRI recordings, the brain is parcellated into 424 regions, which are connected using a K-nearest neighbors graph based on the 3D spatial coordinates of each brain region.}
  \label{fig:graph_creation_fig}
\end{figure}

\subsection{Node tokenization}

To unify the node tokenization procedure for GNNs with that of pretrained transformers, which tokenize inputs into sequences of feature tokens, we define transformation $\tau$ as a function which takes as input node features \(X_i \in \mathbb{R}^{f \times c}\), where \(f\) is the number of features per node and \(c\) is the dimensionality of each feature, and outputs a sequence of $f$ $d$-dimensional feature vectors which will represent node $i$. A general formulation of $\tau$ can be written as: 

\begin{equation}\label{eq:node_tokenization}
    H_i = \tau (X_i) = \texttt{CONCAT}(X_i \textbf{W}, P) \in \mathbb{R}^{f \times d}
\end{equation}

where \textbf{W} is a $c \times d$ learned projection into a $d$-dimensional feature vector, and \(P \in \mathbb{R}^{f \times d}\) are positional encodings for each feature.

We note that this general formulation of $\tau$ becomes dataset-specific following tokenization schemes defined by foundation models on different data modalities. For instance, on datasets with input node feature vectors \(\vec{x}_i \in \mathbb{R}^f\), such as a gene expression vector for a cell containing \(f\) genes, we can see \(X_i\) as an expanded feature vector with \(c = 1\), and \textbf{W} as a projection of a scalar gene expression value into a \(d\)-dimensional vector embedding. The positional encoding \(P\) would then represent a learned gene embedding dictionary \(P \in \mathbb{R}^{f \times d}\), analogous to word embeddings in natural language. The concatenation operation in equation \ref{eq:node_tokenization} would then combine the learned projection of each expression value with its corresponding gene positional encoding, as in scGPT \cite{cui2023scgpt} and Geneformer \cite{theodoris2023transfer}.

For experiments on image datasets, $\tau$ is formulated as a patch encoding procedure following standard ViTs \cite{dosovitskiy2020image}, where an input image is divided into \(f\) patches, each with \(c\) pixels, that are embedded via a learned patch projector \textbf{W}. Positional encoding \(P\) is done through fixed 2D sinusoidal positional encoding which is concatenated with each patch embedding. For fMRI brain activity recordings, $\tau$ follows a spatiotemporal patching process as in the BrainLM foundation model \cite{ortega2023brainlm}, where for each brain region, segments of \(c = 20\) signal timepoints are embedded via a learned projection \textbf{W}. Spatial positional encoding is done through a learned projection of XYZ coordinates of each brain region, and temporal positional encoding is done using sinusoidal positional encoding.

\subsection{Message creation}

Our objective is to formulate message creation between two nodes such that pretrained foundation models can be leveraged to create the messages while fitting into the rest of the message-passing framework. Our key observation is that transformer-based foundation models operate using self-attention over sequences of feature tokens (depicted in Figure \ref{fig:FIMP_operator_figure1}), and contain learned attention weights per layer which are trained to highlight important interactions between feature tokens. Message creation between neighboring node feature sequences can be viewed as a problem of highlighting relevant information which source node $j$ must pass to destination node $i$, and thus the pretrained attention weights can be repurposed for message creation between two nodes.

We define a cross-node attention-based message creation module which takes as input node feature sequences $H_i$ and $H_j$, and outputs a message token sequence $H_{ji}$ which will be passed from node $j$ to node $i$:

\begin{equation}\label{eq:FIMP_attn_QKV}
    Q = H_i \textbf{W}_Q, \
    K = H_j \textbf{W}_K, \
    V = H_j \textbf{W}_V, \
\end{equation}

\begin{equation}\label{eq:message_creation}
    H_{ji} = {\texttt{softmax}\left( \frac{Q K^\top}{\sqrt{d}} \right) V}
\end{equation}

where \textbf{W}$_Q$, \textbf{W}$_K$, and \textbf{W}$_V$ are learned weight matrices which parameterize the attention mechanism. Note that the attention weights can be randomly initialized and learned from scratch, or initialized from pretrained attention weights. Messages $H_{ji}$ can then be aggregated and used to complete the regular message passing aggregation and update steps, with each node represented by a sequence of feature tokens rather than a single embedding vector. The full algorithm is detailed in Algorithm \ref{alg:FIMP_algorithm}.

We note that the cross-attention-based message passing operation in FIMP is fundamentally different from other attention-based GNNs. FIMP is the first method that uses feature-based cross-node attention to construct messages for message passing on graphs. In contrast, attention-based GNNs, particularly GATs and Graph Transformers, do node-level attention and learn scalar attention coefficients between nodes. An overview of attention-based GNNs is provided in the Related Works (section \ref{sec:related_works}), along with a summary of key differences with FIMP.

\begin{algorithm}
\caption{FIMP}
\label{alg:FIMP_algorithm}
\begin{algorithmic}  
\REQUIRE Graph \( G = ( V, E)\), input features \( X_i \in \mathbb{R}^{f \times c} \)
    \STATE $H_i^0 \gets \tau(X_i)$
    \FOR{$k = 1...K$}
        \FOR{node $i \in V$}
            \FOR{node $j \in \mathcal N(i)$}
                \STATE $Q=H_i^{(k-1)} \textbf{W}_Q$
                \STATE $K=H_j^{(k-1)} \textbf{W}_K$
                \STATE $V=H_j^{(k-1)} \textbf{W}_V$
                \STATE $H_{ji}^{(k)} = {\texttt{softmax}\left( \frac{Q K^\top}{\sqrt{d}} \right) V}$
            \ENDFOR
            \STATE $H_{\mathcal N(i)}^{(k)} = \underset{j \in \mathcal{N}(i)}{\texttt{AGGREGATE}}\left( H_{ji}^{(k)} \right)$
            \STATE \( H_i^{(k)} = \texttt{CONCAT}(H_i^{(k-1)}, H_{\mathcal N(i)}^{(k)}) \textbf{W} \)  
        \ENDFOR
    \ENDFOR
\end{algorithmic}
\end{algorithm}

\subsection{Leveraging non-textual foundation models}

In its base formulation, cross-attention message passing can be done with a simple cross-attention mechanism which is learned from scratch during training. We denote this base version of our architecture as FIMP-base in our experiments. Pretrained foundation models, however, can be repurposed to do the message creation in order to leverage their pretraining over vast amounts of unstructured data. Given a pretrained foundation model $\mathcal{F}$ with learned attention weights per each transformer layer, we adapt the self-attention mechanism in each layer to do cross attention between node feature sequences from neighboring nodes. This adaptation is done in each layer by using the pretrained \textbf{W}$_Q$, \textbf{W}$_K$, and \textbf{W}$_V$ weights to project both the source and destination node feature sequences $H_j$ and $H_i$, and computing the scaled dot product attention outlined in equation \ref{eq:message_creation}. The final hidden representation output of the foundation model is then taken as the message $H_{ji}$.

\begin{figure}
  \centering
  \includegraphics[width=1.0\linewidth]{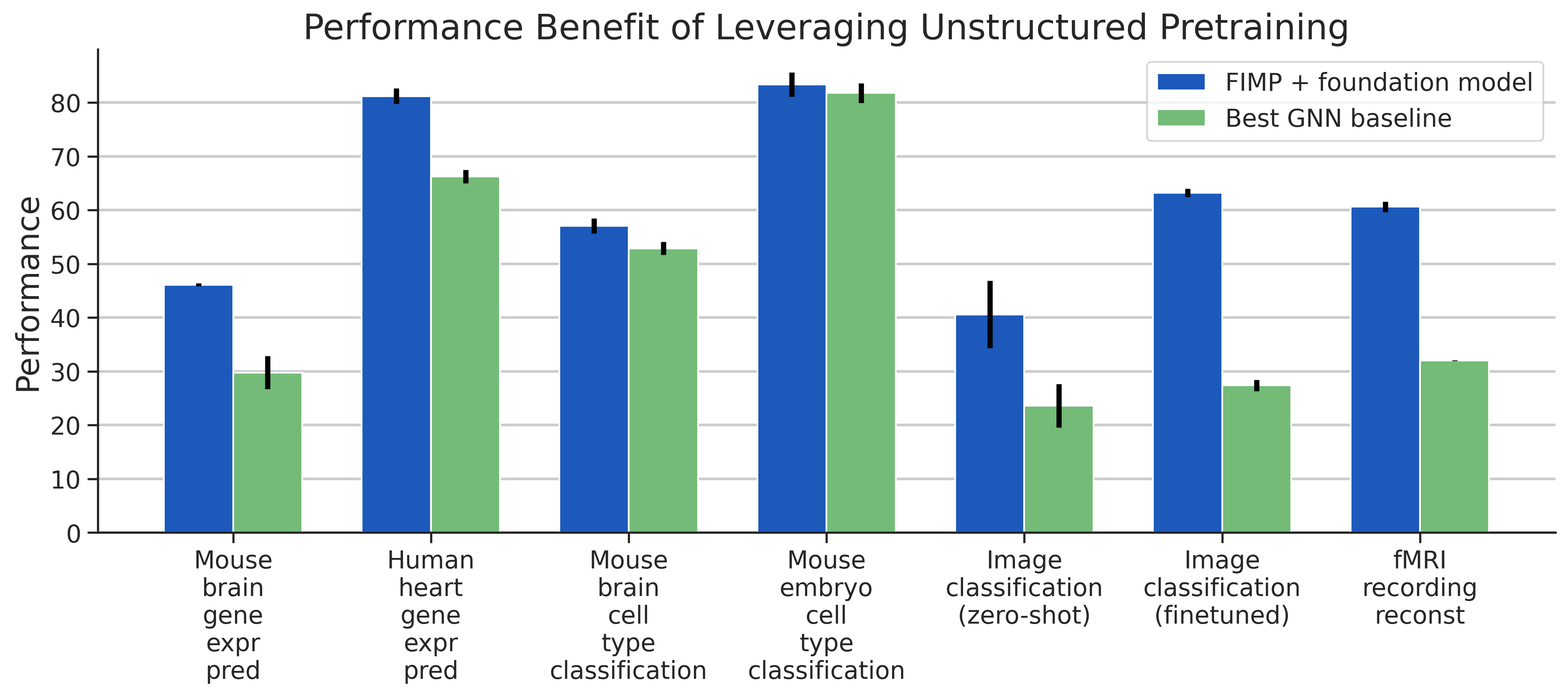}
  \caption{Performance summary across different tasks for FIMP + foundation model versus the best traditional GNN baseline. FIMP improves over traditional GNNs across multiple datasets, highlighting the benefits of leveraging foundation models pretrained on unstructured data.}
  \label{fig:unstructured_pretraining_benefit}
\end{figure}

\section{Experiments}
\label{sec:experiments}

In this section, we demonstrate the effectiveness of our proposed framework on a diverse range of tasks in graph-structured settings: (i) gene expression reconstruction and cell type classification on spatial transcriptomics datasets, (ii) image classification on the Mapillary street-view image dataset, and (iii) brain activity reconstruction on fMRI brain recordings from the UK Biobank (UKB) dataset \cite{miller2016multimodal}. The graph structure inherent in each of these datasets is depicted in Figure \ref{fig:graph_creation_fig}. Our main goal is to show that FIMP allows for the effective integration of pretrained non-textual foundation models into a message-passing framework on graphs.

\subsection{Datasets}

\textbf{Spatial transcriptomics.} We benchmark FIMP on gene expression prediction and cell type classification using three publicly-available spatial transcriptomics datasets. The Slideseq-V2 spatial transcriptomics dataset \cite{stickels2021highly} is a mouse hippocampus dataset consisting of \(41,786\) cells and \(4,000\) genes, with \(14\) different cell types. A second spatial dataset of human heart tissue was obtained from the 10X Genomics public spatial data repository, consisting of \(4,247\) cells each with \(36,601\) measured genes. A third spatial dataset, SeqFISH \cite{lohoff2020highly}, consists of \(15,000\) cells and \(342\) genes taken from mouse embryo tissue sections. For all spatial transcriptomics datasets, we follow standard preprocessing and normalization procedures for RNA sequencing data, including count normalization and log transformation \cite{haque2017practical}. Full dataset details are in Appendix section \ref{appendix:datasets}.

\textbf{Mapillary image dataset.} The Mapillary planet-scale image dataset \cite{antequera2020mapillary} is a dataset of 750,000 street-view images collected from over 170 countries around the world. Images are 1000-2000 pixels in height and width, originating from a variety of cameras and conditions depicting natural landscapes and buildings. Each image has a recorded latitude and longitude coordinate, forming a geographical proximity graph where each node represents a full image, connected to nearby image nodes if they are within 10 miles of one another. We evaluate FIMP on a task where the aim is to classify the country of origin based on the visual features of each image node and its neighborhood. We train on 100,000 training images, and test on the predefined 10,000 test image set, with country labels determined for each image based on its latitude and longitude coordinates.

\textbf{fMRI brain activity recordings.} The UK Biobank dataset \cite{miller2016multimodal} comprises of 76,296 task-based and resting-state functional MRI (fMRI) recordings from 41,986 patients aged 40 to 69 years old. All recordings went through standard preprocessing steps for fMRI recordings \cite{salimi2014automatic, abdallah2021brain}, and was parcellated into 424 brain regions using the AAL-424 atlas \cite{nemati2020unique}. We apply robust scaling per brain region by subtracting the median and dividing by the interquartile range computed across subjects. Our training set comprised of \(60,000\) recordings, with the rest reserved for validation and test.

\begin{table}  
  \caption{Gene expression prediction results on the mouse hippocampus and human heart spatial transcriptomics datasets. Performance is reported across 5 runs in terms of MSE and $R^2$. FIMP outperforms baseline methods on predicting gene expression on both datasets.}
  \label{table:spatial_transcriptomics_table}
  \begin{center}
  \begin{adjustbox}{width=1\textwidth}
  \begin{tabular}{l cc cc}
    \toprule
    \multirow{2}{*}{\makebox[6em]{Method}}  & \multicolumn{2}{c}{Mouse Hippocampus} & \multicolumn{2}{c}{Human Heart} \\
& MSE $(\downarrow)$  & $R^2 (\uparrow)$  & MSE $(\downarrow)$  & $R^2 (\uparrow)$ \\
    \midrule

    GCN         & 0.0211 \(\pm\) 0.0018  & 0.0236 \(\pm\) 0.0457  & 0.0045 \(\pm\) 0.00019  & 0.3368 \(\pm\) 0.04453  \\
    GraphSAGE   & 0.0181 \(\pm\) 0.0012  & 0.1853 \(\pm\) 0.0306  & 0.0054 \(\pm\) 0.00033  & 0.2080 \(\pm\) 0.01973  \\
    GAT         & 0.0201 \(\pm\) 0.0008  & 0.0905 \(\pm\) 0.0233  & 0.0043 \(\pm\) 0.00023  & 0.3468 \(\pm\) 0.02313  \\
    GIN         & 0.0175 \(\pm\) 0.0009  & 0.1707 \(\pm\) 0.0424  & 0.0025 \(\pm\) 0.00029  & 0.6625 \(\pm\) 0.01269  \\
    GraphMAE    & 0.0178 \(\pm\) 0.0006  & 0.1538 \(\pm\) 0.0254  & 0.0024 \(\pm\) 0.00016  & 0.6589 \(\pm\) 0.01715  \\
    GPS         & 0.0149 \(\pm\) 0.0012  & 0.2977 \(\pm\) 0.0308  & 0.0024 \(\pm\) 0.00031  & 0.6538 \(\pm\) 0.01043  \\
    FIMP-base (ours)      & \textit{0.0134 \(\pm\) 0.0009}  & \textit{0.3815 \(\pm\) 0.0226}  &  \textit{0.0021 \(\pm\) 0.00003}  & \textit{0.6955 \(\pm\) 0.02048}  \\
    FIMP + ViT (ours)   &  0.0128 \(\pm\) 0.0010  &  0.3506 \(\pm\) 0.0452  & 0.0042 \(\pm\) 0.00089  & 0.4026 \(\pm\) 0.08102  \\
    FIMP + GenePT (ours)   &  \underline{0.0129 \(\pm\) 0.0005}  &  \underline{0.4058 \(\pm\) 0.0302}  & \underline{0.0013 \(\pm\) 0.00023}  & \underline{0.7952 \(\pm\) 0.01430}  \\
    FIMP + scGPT (ours)    & \textbf{0.0119 \(\pm\) 0.0008}  & \textbf{0.4612 \(\pm\) 0.0029}  & \textbf{0.0011 \(\pm\) 0.00008}  & \textbf{0.8119 \(\pm\) 0.01428}  \\
    \bottomrule
  \end{tabular}
  \end{adjustbox}
  \end{center}
\end{table}

\begin{table}  
  \caption{Cell type classification results on the mouse hippocampus and embryo spatial transcriptomics datasets. Performance is reported in terms of accuracy and F1-score. FIMP outperforms baseline models at predicting cell types.}
  \label{table:cell_type_image_classification_table}
  \centering
  \begin{tabular}{l c c c c}
    \toprule
    & \multicolumn{2}{c}{Mouse Hippocampus} & \multicolumn{2}{c}{Mouse Embryo} \\
    Method  & Accuracy ($\uparrow$) & F1-score ($\uparrow$) & Accuracy ($\uparrow$) & F1-score ($\uparrow$)  \\
    \midrule
    GCN         & 47.59 \(\pm\) 3.788  & 0.445 \(\pm\) 0.050 & 74.23 \(\pm\) 1.250 & 0.720 \(\pm\) 0.008  \\
    GraphSAGE   & 51.81 \(\pm\) 3.229  & 0.495 \(\pm\) 0.036 & 80.77 \(\pm\) 3.071 & 0.793 \(\pm\) 0.031  \\
    GAT         & 46.21 \(\pm\) 3.110  & 0.442 \(\pm\) 0.031 & 71.07 \(\pm\) 1.452 & 0.690 \(\pm\) 0.014  \\
    GIN         & 52.71 \(\pm\) 0.421  & 0.507 \(\pm\) 0.008 & 75.51 \(\pm\) 1.398 & 0.743 \(\pm\) 0.012  \\
    GPS         & \textit{52.89 \(\pm\) 1.176}  & \textit{0.510 \(\pm\) 0.008} & \textit{81.77 \(\pm\) 3.175} & \textit{0.813 \(\pm\) 0.038}  \\
    FIMP-base   & 49.04 \(\pm\) 1.215  & 0.464 \(\pm\) 0.019 & 81.35 \(\pm\) 2.285 & 0.807 \(\pm\) 0.026  \\
    scGPT       & \underline{53.50 \(\pm\) 0.424} & \underline{0.518 \(\pm\) 0.005}  & \underline{82.93 \(\pm\) 0.419} & \underline{0.820 \(\pm\) 0.005}  \\
    FIMP-scGPT  & \textbf{57.05 \(\pm\) 1.393}  & \textbf{0.554 \(\pm\) 0.004}  & \textbf{83.33 \(\pm\) 2.250} & \textbf{0.821 \(\pm\) 0.022} \\
    \bottomrule
  \end{tabular}
\end{table}

\subsection{Experimental setup}

All models were implemented in Pytorch Geometric \cite{fey2019fast} and Pytorch \cite{paszke2019pytorch}, and trained using the Adam optimizer \cite{kingma2014adam}. Flash Attention \cite{dao2022flashattention} is used to improve the computational footprint during message passing. Hyperparameter tuning was done through a grid search over standard values for learning rate, dropout, attention dropout, and weight decay. For all experiments, a 24GB NVIDIA GPU (RTX3090 or A5000) was used for training. Experimental setup details for specific datasets are provided in the Appendix \ref{appendix:expt_setup} due to space limitations.

\textbf{Foundation models}. For experiments on single-cell datasets, the scGPT \cite{cui2023scgpt} whole-human checkpoint is incorporated for message creation in FIMP-scGPT, consisting of a 12-layer transformer with 54 million parameters. scGPT is pretrained using a masked gene expression prediction objective on over 33 million cells from a diverse array of human tissues and organs. The pretrained gene embedding table is also utilized from the pretrained scGPT checkpoint, representing pretrained knowledge about gene identities in transcriptomics datasets. Additionally, we also utilize the gene embeddings obtained by GenePT \cite{chen2023genept}, which are GPT-3.5 embeddings of gene function descriptions based on biomedical literature, as another pretrained gene embedding experiment. For image classification, a standard ViT \cite{dosovitskiy2020image} with 12 transformer layers and 86 million parameters is used as a message creator. The patch encoder from the ViT is also reused from the ViT embedding module. For experiments on fMRI brain recordings, the BrainLM \cite{ortega2023brainlm} model was used, which consists of a Masked Autoencoder transformer with an 8-layer encoder and 4-layer decoder, totaling 26 million parameters.

\textbf{Baselines}. For both supervised and self-supervised tasks, we compare FIMP against popular message-passing GNN architectures, including GCN \cite{kipf2016semi}, GraphSAGE \cite{hamilton2017inductive}, Graph Attention Networks (GATs) \cite{velivckovic2017graph}, and Graph Isomorphism Networks (GINs) \cite{xu2018powerful}. We also compare FIMP against more recent GNN architectures, namely GraphMAE \cite{hou2022graphmae}, a masked graph autoencoder model, and GPS Graph Transformer \cite{rampavsek2022recipe}, a SOTA graph transformer framework. For supervised classification tasks, we additionally compare to the pretrained foundation model in each domain, which does not take graph structure as input and instead treats each node as an individual sample.

\begin{table}  
  \caption{Image classification results on the Mapillary street-view image dataset. FIMP significantly improves over baseline models in image classification, and creates zero-shot embeddings of the image network on par with trained GNN baseline models.}
  \label{table:mapillary_image_classification_table}
  \centering
  \begin{tabular}{l l c c}
    \toprule
    Setting  & Method  & Accuracy ($\uparrow$) & F1-score ($\uparrow$)  \\
    \midrule
    \multirow{9}{*}{\makebox[3em]{Finetuned}}
    & GCN         & 23.9 \(\pm\) 1.152  & 0.182 \(\pm\) 0.0151  \\
    & GraphSAGE   & 22.2 \(\pm\) 1.703  & 0.164 \(\pm\) 0.0129  \\
    & GAT         & 22.9 \(\pm\) 0.596  & 0.189 \(\pm\) 0.0042  \\
    & GIN         & 26.4 \(\pm\) 1.240  & 0.254 \(\pm\) 0.0143  \\
    & GraphMAE    & 15.8 \(\pm\) 0.828  & 0.083 \(\pm\) 0.0056  \\
    & GPS         & 27.4 \(\pm\) 1.046  & 0.268 \(\pm\) 0.0157  \\
    & FIMP-base (ours)   & \textit{38.6 \(\pm\) 1.174}  & \textit{0.422 \(\pm\) 0.0170}  \\
    & ViT         & \underline{56.5 \(\pm\) 3.187}      & \underline{0.597 \(\pm\) 0.0065}  \\
    & FIMP-ViT (ours)    & \textbf{63.2 \(\pm\) 0.764}  & \textbf{0.684 \(\pm\) 0.0076}  \\
    \midrule
    \multirow{4}{*}{\makebox[3em]{Zero-shot}}
    & Majority class    & 17.0 \(\pm\) 3.162  & --  \\
    & GraphSAGE         & 23.6 \(\pm\) 4.037  & 0.129 \(\pm\) 0.0309  \\
    & ViT               & 34.0 \(\pm\) 3.391  & 0.282 \(\pm\) 0.0389  \\
    & FIMP-ViT (ours)   & \textbf{40.6 \(\pm\) 6.269}  & \textbf{0.371 \(\pm\) 0.0550}  \\
    \bottomrule
  \end{tabular}
\end{table}

\subsection{Results}

\textbf{Spatial transcriptomics}. Table \ref{table:spatial_transcriptomics_table} contains results for gene expression prediction on the human heart and mouse hippocampus datasets. From these results, we observe that FIMP-base, trained from scratch with a randomly initialized cross-attention layer as a message creator, is able to outperform baseline GNNs at predicting masked gene expression values. We attribute this to improved gene tokenization, with the learned gene embedding table capturing information about different genes from the data. When we leverage pretrained gene embeddings learned on unstructured data, either from GenePT \cite{chen2023genept} or scGPT \cite{cui2023scgpt} (denoted as FIMP-GenePT and FIMP-scGPT, respectively), we observe further increases in gene expression prediction performance. 
Interestingly, we note that using an out-of-domain foundation model such as ViT as the message creator does not improve performance, suggesting that performance improvements are not trivially caused by increased model capacity, and rather depend on the pretraining domain being sufficiently aligned with the graph features.

Table \ref{table:cell_type_image_classification_table} contains results for cell type classification on the mouse hippocampus and embryo spatial transcriptomics datasets. We note that in this supervised classification task, FIMP-scGPT achieves the highest classification performance on both datasets.

\begin{table*}  
  \caption{Brain activity reconstruction results on the UK Biobank dataset. Performance is reported across 5 runs in terms of Mean Squared Error (MSE) and $R^2$. FIMP improves upon baselines by \(25.8\%\), with a further improvement of \(2.8\%\) by leveraging BrainLM \cite{ortega2023brainlm} for message creation.}
  \label{table:fmri_reconstruction_benchmark_table}
  \centering
  \begin{tabular}{llll}
    \toprule
    Method  & Masking Strategy & MSE ($\downarrow$) & $R^2$ ($\uparrow$)  \\
    \midrule
    \multirow{3}{6em}{GCN}
    & Replace noise         & 0.554 \(\pm\) 0.00002  & 0.189 \(\pm\) 0.00003  \\
    & Fill in mean          & 0.513 \(\pm\) 0.00019  & 0.248 \(\pm\) 0.00028  \\
    & Linear interpolation  & 0.535 \(\pm\) 0.00137  & 0.217 \(\pm\) 0.00200  \\
    \midrule
    \multirow{3}{6em}{GraphSAGE} 
    & Replace noise         & 0.534 \(\pm\) 0.00107  & 0.218 \(\pm\) 0.00157  \\
    & Fill in mean          & \textit{0.464 \(\pm\) 0.00039}  & \textit{0.320 \(\pm\) 0.00057}  \\
    & Linear interpolation  & 0.500 \(\pm\) 0.00094  & 0.268 \(\pm\) 0.00138  \\
    \midrule
    \multirow{3}{6em}{GAT} 
    & Replace noise         & 0.548 \(\pm\) 0.00004  & 0.197 \(\pm\) 0.00007  \\
    & Fill in mean          & 0.505 \(\pm\) 0.00005  & 0.260 \(\pm\) 0.00007  \\
    & Linear interpolation  & 0.527 \(\pm\) 0.00052  & 0.229 \(\pm\) 0.00076  \\
    \midrule
    \multirow{3}{6em}{GIN} 
    & Replace noise         & 0.564 \(\pm\) 0.00131  & 0.174 \(\pm\) 0.00192  \\
    & Fill in mean          & 0.533 \(\pm\) 0.00185  & 0.220 \(\pm\) 0.00271  \\
    & Linear interpolation  & 0.559 \(\pm\) 0.00061  & 0.181 \(\pm\) 0.00090  \\
    \midrule
    \multirow{3}{6em}{GraphMAE} 
    & Replace noise         & 0.582 \(\pm\) 0.00070  & 0.147 \(\pm\) 0.00103  \\
    & Fill in mean          & 0.544 \(\pm\) 0.00030  & 0.203 \(\pm\) 0.00044  \\
    & Linear interpolation  & 0.573 \(\pm\) 0.00091  & 0.160 \(\pm\) 0.00134  \\
    \midrule
    \multirow{3}{6em}{GPS Graph Transformer} 
    & Replace noise         & 0.577 \(\pm\) 0.00279  & 0.154 \(\pm\) 0.00408 \\
    & Fill in mean          & 0.547 \(\pm\) 0.01030  & 0.198 \(\pm\) 0.01506  \\
    & Linear interpolation  & 0.557 \(\pm\) 0.01034  & 0.184 \(\pm\) 0.01512  \\
    \midrule
    FIMP-base & Tokenization + PE    & \underline{0.288 \(\pm\) 0.00713}  & \underline{0.578 \(\pm\) 0.01043}  \\
    FIMP-BrainLM & Tokenization + PE  & \textbf{0.267 \(\pm\) 0.00493}  & \textbf{0.606 \(\pm\) 0.00972}  \\
    
    \bottomrule
  \end{tabular}
\end{table*}

\textbf{Image classification}. Table \ref{table:mapillary_image_classification_table} summarizes results for image classification on the Mapillary image dataset. We observe that FIMP-base outperforms baseline GNNs by over \(10\%\) due to its improved tokenization of image patches, despite being learned from scratch. The best performance is obtained by FIMP-ViT, which utilizes a pretrained ViT \cite{dosovitskiy2020image} for cross-node message creation. 

\textbf{Zero-shot node embedding}. We furthermore explore a zero-shot setting for embedding image networks, to evaluate the capability of FIMP to leverage the pretrained ViT model without any graph-specific training. We embed subgraphs of the Mapillary dataset with FIMP, and compare it to embeddings generated by a randomly initialized GraphSAGE model \cite{hamilton2017inductive} and the ViT model itself with no graph structure, with 400 image embeddings obtained per model. We evaluate the quality of embeddings by training a linear classifier on \(75\%\) of the embeddings and predicting labels for the remaining \(25\%\). We observe that FIMP-ViT is able to generate zero-shot embeddings which get over \(40\%\) classification accuracy, on par with finetuned baseline GNNs despite having no graph-specific training. This strongly indicates that FIMP is able to effectively leverage pretrained non-textual foundation models, and enables exciting zero-shot application scenarios which were previously not possible with non-textual foundation models operating on unstructured data.

\textbf{fMRI recording reconstruction}. Table \ref{table:fmri_reconstruction_benchmark_table} summarizes results for fMRI recording reconstruction on the UK Biobank \cite{miller2016multimodal} dataset. FIMP-base improves upon baseline GNNs by \(25\%\) in terms of reconstruction performance on masked brain signals, with a further performance improvement of around \(3\%\) from leveraging the pretrained BrainLM \cite{ortega2023brainlm} model for cross-node message creation.

\section{Conclusions, Limitations, and Future Research}
\label{sec:conclusion}

In this work, we introduce Foundation-Informed Message Passing (FIMP), a message-passing framework which repurposes the self-attention layers of pretrained non-textual foundation models for message-passing on graphs. Our approach represents the first broad exploration of utilizing non-textual foundation models pretrained on vast amounts unstructured data in graph-based tasks. FIMP demonstrates improved performance over baselines across multiple tasks in image networks, spatial transcriptomics data, and fMRI brain activity recordings, confirming the performance benefits of leveraging non-textual foundation models in graph-based tasks. Furthermore, FIMP demonstrates zero-shot embedding capabilities on image networks that are on par with trained GNNs. This enables zero-shot applications on graphs, previously impossible with non-textual foundation models.

There are several avenues for improvement upon our method, which we leave for future work. Currently, our evaluation of FIMP is limited to image and biological data. Protein design and social networks are promising areas of future research. Additionally, extending FIMP to a more general pretraining scheme for graph foundation models, as well as supporting multimodal graphs, heterogeneous graphs, and edge features, would all expand the potential applications of FIMP. Finally, improving the computational footprint of FIMP through strategies such as feature selection and efficient attention mechanisms beyond Flash Attention is an important future direction.




\bibliographystyle{plain} 
\bibliography{main} 

\begin{thebibliography}{10}

\bibitem{aaberg2018co}
Charlotte Aaberg-Jessen, Mia~D S{\o}rensen, Ana~LSA Matos, Jos{\'e}~M Moreira, Nils Br{\"u}nner, Arnon Knudsen, and Bjarne~W Kristensen.
\newblock Co-expression of timp-1 and its cell surface binding partner cd63 in glioblastomas.
\newblock {\em BMC cancer}, 18:1--16, 2018.

\bibitem{abdallah2021brain}
Chadi~G Abdallah.
\newblock Brain networks associated with covid-19 risk: Data from 3662 participants.
\newblock {\em Chronic Stress}, 5:24705470211066770, 2021.

\bibitem{antequera2020mapillary}
Manuel~L{\'o}pez Antequera, Pau Gargallo, Markus Hofinger, Samuel~Rota Bul{\`o}, Yubin Kuang, and Peter Kontschieder.
\newblock Mapillary planet-scale depth dataset.
\newblock In {\em Computer Vision--ECCV 2020: 16th European Conference, Glasgow, UK, August 23--28, 2020, Proceedings, Part II 16}, pages 589--604. Springer, 2020.

\bibitem{bommasani2021opportunities}
Rishi Bommasani, Drew~A Hudson, Ehsan Adeli, Russ Altman, Simran Arora, Sydney von Arx, Michael~S Bernstein, Jeannette Bohg, Antoine Bosselut, Emma Brunskill, et~al.
\newblock On the opportunities and risks of foundation models.
\newblock {\em arXiv preprint arXiv:2108.07258}, 2021.

\bibitem{brown2020language}
Tom Brown, Benjamin Mann, Nick Ryder, Melanie Subbiah, Jared~D Kaplan, Prafulla Dhariwal, Arvind Neelakantan, Pranav Shyam, Girish Sastry, Amanda Askell, et~al.
\newblock Language models are few-shot learners.
\newblock {\em Advances in neural information processing systems}, 33:1877--1901, 2020.

\bibitem{chen2020uniter}
Yen-Chun Chen, Linjie Li, Licheng Yu, Ahmed El~Kholy, Faisal Ahmed, Zhe Gan, Yu~Cheng, and Jingjing Liu.
\newblock Uniter: Universal image-text representation learning.
\newblock In {\em European conference on computer vision}, pages 104--120. Springer, 2020.

\bibitem{chen2023genept}
Yiqun~T Chen and James Zou.
\newblock Genept: A simple but hard-to-beat foundation model for genes and cells built from chatgpt.
\newblock {\em bioRxiv}, pages 2023--10, 2023.

\bibitem{cui2023scgpt}
Haotian Cui, Chloe Wang, Hassaan Maan, Kuan Pang, Fengning Luo, and Bo~Wang.
\newblock scgpt: towards building a foundation model for single-cell multi-omics using generative ai.
\newblock {\em bioRxiv}, pages 2023--04, 2023.

\bibitem{dao2022flashattention}
Tri Dao, Dan Fu, Stefano Ermon, Atri Rudra, and Christopher R{\'e}.
\newblock Flashattention: Fast and memory-efficient exact attention with io-awareness.
\newblock {\em Advances in Neural Information Processing Systems}, 35:16344--16359, 2022.

\bibitem{devlin2018bert}
Jacob Devlin, Ming-Wei Chang, Kenton Lee, and Kristina Toutanova.
\newblock Bert: Pre-training of deep bidirectional transformers for language understanding.
\newblock {\em arXiv preprint arXiv:1810.04805}, 2018.

\bibitem{dey2023revisiting}
Saurabh Dey, Soumya Basu, and Amit Ranjan.
\newblock Revisiting the role of cd63 as pro-tumorigenic or anti-tumorigenic tetraspanin in cancers and its theragnostic implications.
\newblock {\em Advanced Biology}, page 2300078, 2023.

\bibitem{dosovitskiy2020image}
Alexey Dosovitskiy, Lucas Beyer, Alexander Kolesnikov, Dirk Weissenborn, Xiaohua Zhai, Thomas Unterthiner, Mostafa Dehghani, Matthias Minderer, Georg Heigold, Sylvain Gelly, et~al.
\newblock An image is worth 16x16 words: Transformers for image recognition at scale.
\newblock {\em arXiv preprint arXiv:2010.11929}, 2020.

\bibitem{dwivedi2020generalization}
Vijay~Prakash Dwivedi and Xavier Bresson.
\newblock A generalization of transformer networks to graphs.
\newblock {\em arXiv preprint arXiv:2012.09699}, 2020.

\bibitem{fatemi2023talk}
Bahare Fatemi, Jonathan Halcrow, and Bryan Perozzi.
\newblock Talk like a graph: Encoding graphs for large language models.
\newblock {\em arXiv preprint arXiv:2310.04560}, 2023.

\bibitem{fey2019fast}
Matthias Fey and Jan~Eric Lenssen.
\newblock Fast graph representation learning with pytorch geometric.
\newblock {\em arXiv preprint arXiv:1903.02428}, 2019.

\bibitem{gilmer2017neural}
Justin Gilmer, Samuel~S Schoenholz, Patrick~F Riley, Oriol Vinyals, and George~E Dahl.
\newblock Neural message passing for quantum chemistry.
\newblock In {\em International conference on machine learning}, pages 1263--1272. PMLR, 2017.

\bibitem{guo2023gpt4graph}
Jiayan Guo, Lun Du, and Hengyu Liu.
\newblock Gpt4graph: Can large language models understand graph structured data? an empirical evaluation and benchmarking.
\newblock {\em arXiv preprint arXiv:2305.15066}, 2023.

\bibitem{gururangan2020don}
Suchin Gururangan, Ana Marasovi{\'c}, Swabha Swayamdipta, Kyle Lo, Iz~Beltagy, Doug Downey, and Noah~A Smith.
\newblock Don't stop pretraining: Adapt language models to domains and tasks.
\newblock {\em arXiv preprint arXiv:2004.10964}, 2020.

\bibitem{hamilton2017inductive}
Will Hamilton, Zhitao Ying, and Jure Leskovec.
\newblock Inductive representation learning on large graphs.
\newblock {\em Advances in neural information processing systems}, 30, 2017.

\bibitem{haque2017practical}
Ashraful Haque, Jessica Engel, Sarah~A Teichmann, and Tapio L{\"o}nnberg.
\newblock A practical guide to single-cell rna-sequencing for biomedical research and clinical applications.
\newblock {\em Genome medicine}, 9:1--12, 2017.

\bibitem{he2022masked}
Kaiming He, Xinlei Chen, Saining Xie, Yanghao Li, Piotr Doll{\'a}r, and Ross Girshick.
\newblock Masked autoencoders are scalable vision learners.
\newblock In {\em Proceedings of the IEEE/CVF conference on computer vision and pattern recognition}, pages 16000--16009, 2022.

\bibitem{hou2022graphmae}
Zhenyu Hou, Xiao Liu, Yukuo Cen, Yuxiao Dong, Hongxia Yang, Chunjie Wang, and Jie Tang.
\newblock Graphmae: Self-supervised masked graph autoencoders.
\newblock In {\em Proceedings of the 28th ACM SIGKDD Conference on Knowledge Discovery and Data Mining}, pages 594--604, 2022.

\bibitem{kearney2022brain}
Breanne~E Kearney and Ruth~A Lanius.
\newblock The brain-body disconnect: A somatic sensory basis for trauma-related disorders.
\newblock {\em Frontiers in Neuroscience}, 16:1881, 2022.

\bibitem{kingma2014adam}
Diederik~P Kingma and Jimmy Ba.
\newblock Adam: A method for stochastic optimization.
\newblock {\em arXiv preprint arXiv:1412.6980}, 2014.

\bibitem{kipf2016semi}
Thomas~N Kipf and Max Welling.
\newblock Semi-supervised classification with graph convolutional networks.
\newblock {\em arXiv preprint arXiv:1609.02907}, 2016.

\bibitem{kreuzer2021rethinking}
Devin Kreuzer, Dominique Beaini, Will Hamilton, Vincent L{\'e}tourneau, and Prudencio Tossou.
\newblock Rethinking graph transformers with spectral attention.
\newblock {\em Advances in Neural Information Processing Systems}, 34:21618--21629, 2021.

\bibitem{liu2023one}
Hao Liu, Jiarui Feng, Lecheng Kong, Ningyue Liang, Dacheng Tao, Yixin Chen, and Muhan Zhang.
\newblock One for all: Towards training one graph model for all classification tasks.
\newblock {\em arXiv preprint arXiv:2310.00149}, 2023.

\bibitem{lohoff2020highly}
Tim Lohoff, Shila Ghazanfar, Alsu Missarova, Noushin Koulena, Nico Pierson, Jonathan~A Griffiths, Evan~S Bardot, C-HL Eng, Richard~CV Tyser, Ricard Argelaguet, et~al.
\newblock Highly multiplexed spatially resolved gene expression profiling of mouse organogenesis.
\newblock {\em BioRxiv}, pages 2020--11, 2020.

\bibitem{miller2016multimodal}
Karla~L Miller, Fidel Alfaro-Almagro, Neal~K Bangerter, David~L Thomas, Essa Yacoub, Junqian Xu, Andreas~J Bartsch, Saad Jbabdi, Stamatios~N Sotiropoulos, Jesper~LR Andersson, et~al.
\newblock Multimodal population brain imaging in the uk biobank prospective epidemiological study.
\newblock {\em Nature neuroscience}, 19(11):1523--1536, 2016.

\bibitem{nemati2020unique}
Samaneh Nemati, Teddy~J Akiki, Jeremy Roscoe, Yumeng Ju, Christopher~L Averill, Samar Fouda, Arpan Dutta, Shane McKie, John~H Krystal, JF~William Deakin, et~al.
\newblock A unique brain connectome fingerprint predates and predicts response to antidepressants.
\newblock {\em IScience}, 23(1), 2020.

\bibitem{ortega2023brainlm}
Josue Ortega~Caro, Antonio~Henrique Oliveira~Fonseca, Christopher Averill, Syed~A Rizvi, Matteo Rosati, James~L Cross, Prateek Mittal, Emanuele Zappala, Daniel Levine, Rahul~M Dhodapkar, et~al.
\newblock Brainlm: A foundation model for brain activity recordings.
\newblock {\em bioRxiv}, pages 2023--09, 2023.

\bibitem{paszke2019pytorch}
Adam Paszke, Sam Gross, Francisco Massa, Adam Lerer, James Bradbury, Gregory Chanan, Trevor Killeen, Zeming Lin, Natalia Gimelshein, Luca Antiga, et~al.
\newblock Pytorch: An imperative style, high-performance deep learning library.
\newblock {\em Advances in neural information processing systems}, 32, 2019.

\bibitem{radford2021learning}
Alec Radford, Jong~Wook Kim, Chris Hallacy, Aditya Ramesh, Gabriel Goh, Sandhini Agarwal, Girish Sastry, Amanda Askell, Pamela Mishkin, Jack Clark, et~al.
\newblock Learning transferable visual models from natural language supervision.
\newblock In {\em International conference on machine learning}, pages 8748--8763. PMLR, 2021.

\bibitem{radford2018improving}
Alec Radford, Karthik Narasimhan, Tim Salimans, Ilya Sutskever, et~al.
\newblock Improving language understanding by generative pre-training.
\newblock 2018.

\bibitem{rampavsek2022recipe}
Ladislav Ramp{\'a}{\v{s}}ek, Michael Galkin, Vijay~Prakash Dwivedi, Anh~Tuan Luu, Guy Wolf, and Dominique Beaini.
\newblock Recipe for a general, powerful, scalable graph transformer.
\newblock {\em Advances in Neural Information Processing Systems}, 35:14501--14515, 2022.

\bibitem{salimi2014automatic}
Gholamreza Salimi-Khorshidi, Gwena{\"e}lle Douaud, Christian~F Beckmann, Matthew~F Glasser, Ludovica Griffanti, and Stephen~M Smith.
\newblock Automatic denoising of functional mri data: combining independent component analysis and hierarchical fusion of classifiers.
\newblock {\em Neuroimage}, 90:449--468, 2014.

\bibitem{stickels2021highly}
Robert~R Stickels, Evan Murray, Pawan Kumar, Jilong Li, Jamie~L Marshall, Daniela~J Di~Bella, Paola Arlotta, Evan~Z Macosko, and Fei Chen.
\newblock Highly sensitive spatial transcriptomics at near-cellular resolution with slide-seqv2.
\newblock {\em Nature biotechnology}, 39(3):313--319, 2021.

\bibitem{svensson2018exponential}
Valentine Svensson, Roser Vento-Tormo, and Sarah~A Teichmann.
\newblock Exponential scaling of single-cell rna-seq in the past decade.
\newblock {\em Nature protocols}, 13(4):599--604, 2018.

\bibitem{theodoris2023transfer}
Christina~V Theodoris, Ling Xiao, Anant Chopra, Mark~D Chaffin, Zeina~R Al~Sayed, Matthew~C Hill, Helene Mantineo, Elizabeth~M Brydon, Zexian Zeng, X~Shirley Liu, et~al.
\newblock Transfer learning enables predictions in network biology.
\newblock {\em Nature}, pages 1--9, 2023.

\bibitem{velivckovic2017graph}
Petar Veli{\v{c}}kovi{\'c}, Guillem Cucurull, Arantxa Casanova, Adriana Romero, Pietro Lio, and Yoshua Bengio.
\newblock Graph attention networks.
\newblock {\em arXiv preprint arXiv:1710.10903}, 2017.

\bibitem{wang2023can}
Heng Wang, Shangbin Feng, Tianxing He, Zhaoxuan Tan, Xiaochuang Han, and Yulia Tsvetkov.
\newblock Can language models solve graph problems in natural language?
\newblock {\em arXiv preprint arXiv:2305.10037}, 2023.

\bibitem{xu2018powerful}
Keyulu Xu, Weihua Hu, Jure Leskovec, and Stefanie Jegelka.
\newblock How powerful are graph neural networks?
\newblock {\em arXiv preprint arXiv:1810.00826}, 2018.

\bibitem{yan2023structural}
Shizhen Yan, Juntao Chen, Xiaojuan Yin, Ziliang Zhu, Ziping Liang, Hua Jin, Han Li, Jianzhong Yin, Yunpeng Jiang, and Yaoyuan Xia.
\newblock The structural basis of age-related decline in global motion perception at fast and slow speeds.
\newblock {\em Neuropsychologia}, 183:108507, 2023.

\bibitem{ying2021transformers}
Chengxuan Ying, Tianle Cai, Shengjie Luo, Shuxin Zheng, Guolin Ke, Di~He, Yanming Shen, and Tie-Yan Liu.
\newblock Do transformers really perform badly for graph representation?
\newblock {\em Advances in neural information processing systems}, 34:28877--28888, 2021.

\bibitem{yuan2021florence}
Lu~Yuan, Dongdong Chen, Yi-Ling Chen, Noel Codella, Xiyang Dai, Jianfeng Gao, Houdong Hu, Xuedong Huang, Boxin Li, Chunyuan Li, et~al.
\newblock Florence: A new foundation model for computer vision.
\newblock {\em arXiv preprint arXiv:2111.11432}, 2021.

\bibitem{yun2019graph}
Seongjun Yun, Minbyul Jeong, Raehyun Kim, Jaewoo Kang, and Hyunwoo~J Kim.
\newblock Graph transformer networks.
\newblock {\em Advances in neural information processing systems}, 32, 2019.

\bibitem{zhang2020graph}
Jiawei Zhang, Haopeng Zhang, Congying Xia, and Li~Sun.
\newblock Graph-bert: Only attention is needed for learning graph representations.
\newblock {\em arXiv preprint arXiv:2001.05140}, 2020.

\end{thebibliography}


\appendix


\section{Datasets (Extended)}
\label{appendix:datasets}

\textbf{Spatial transcriptomics.} We use three publicly-available spatial transcriptomics datasets. The Slideseq-V2 spatial transcriptomics dataset \cite{stickels2021highly} is a mouse hippocampus dataset consisting of \(41,786\) cells and \(4,000\) genes, with \(14\) different cell type classes. A second spatial dataset of human heart tissue was obtained from the 10X Genomics public spatial data repository, consisting of \(4247\) cells each with \(36601\) measured genes. A third spatial dataset, SeqFISH \cite{lohoff2020highly}, consists of \(15,000\) cells and \(342\) genes taken from mouse embryo tissue sections. For all spatial transcriptomics datasets, we follow standard preprocessing and normalization procedures for RNA sequencing data, including count normalization and log transformation \cite{haque2017practical}. For all datasets, we take the intersection of gene features which are present in the scGPT \cite{cui2023scgpt} pretrained foundation model, and split nodes into training, validation, and test sets with a 70/10/20 split. For graph adjacency information, we utilize the neighbor connectivity matrix present in each spatial transcriptomics dataset, which is derived from the original tissue section coordinates.

\textbf{Mapillary image dataset.} The Mapillary planet-scale image dataset \cite{antequera2020mapillary} is a dataset of 750,000 street-view images collected from over 170 countries around the world. Images are 1000-2000 pixels in height and width, originating from a variety of cameras and conditions depicting natural landscapes and buildings. Each image has a recorded latitude and longitude coordinate, forming a geographical proximity graph where each node represents a full image, connected to nearby image nodes if they are within 10 miles of one another. We evaluate FIMP on a geoguesser task, where the aim is to classify the country of origin based on the visual features of each image node and its neighborhood of nearby images. We train on 100,000 training images, and test on the predefined 10,000 test image set, with country labels determined for each image based on its latitude and longitude coordinates.

\textbf{fMRI brain activity recordings.} The UK Biobank dataset \cite{miller2016multimodal} comprises of 76,296 task-based and resting-state functional MRI (fMRI) recordings from 41,986 patients aged 40 to 69 years old. Recordings were acquired on
a Siemens 3T scanner at 0.735s temporal resolution. All recordings went through standard preprocessing steps, including motion correction, normalization, temporal filtering, and ICA denoising \cite{salimi2014automatic, abdallah2021brain}. We parcellated the brain into 424 brain regions using the AAL-424 atlas \cite{nemati2020unique}, yielding 424-dimensional scan sequences sampled at ª1 Hz. Finally, robust scaling was applied by subtracting the median and dividing by the interquartile range computed across subjects for each brain region. Our training set comprised of \(60,000\) of the fMRI recordings, with the rest reserved for validation and test sets.

\section{Experimental Setup (Extended)}  
\label{appendix:expt_setup}

The following section gives additional details about experimental setup across different datasets.

\subsection{Image classification}

For image classification experiments, random 512x512 crops were taken from each image during training, with a 512x512 center crop taken at test time. Per-channel normalization was done on each image using statistics calculated across training images in the Mapillary image dataset. For FIMP and FIMP-ViT experiments, images were divided into 32x32 patches following the standard ViT patch encoding procedure \cite{dosovitskiy2020image}. For baseline GNNs, pixel values for each image were flattened and encoded using a learned projection.

\subsection{Gene expression prediction}

For gene expression prediction experiments on spatial transcriptomics datasets, we limit the number of cells in each dataset to 5\% of the original dataset size, leaving \(1000\) cells for the mouse hippocampus spatial dataset, and \(200\) cells for the human heart spatial dataset. This creates a challenging limited data setting for predicting gene expression values on each spatial dataset. We sample \(50\) nonzero expressed genes in each cell for all models and mask out 80\% of the gene expression values, taking MSE loss against only masked out genes.

\subsection{fMRI recording reconstruction}

In brain activity reconstruction experiments , we sample 320 consecutive timepoints from each fMRI recording, giving a recording of 424 brain regions with 320 timepoints of signal for each region. Each brain region is represented as 1 node in the graph, with node features being the 320 timepoints of signal. We segment the timepoints for each brain region into patches of 20 timepoints, and perform masked reconstruction of brain recording signals. For FIMP and variants of FIMP leveraging foundation models, masked patches are replaced with a mask token, and the signals are predicted back by the model. For baseline GNN models, node features comprise of the 320 timepoints of signal, and we explore three methods for replacing masked out patch values: i) replacing with random noise, ii) filling in with the mean value of the brain region, and iii) linearly interpolating between adjacent non-masked timepoint values. All models mask out $50\%$ of patches per each brain region, with mean squared error (MSE) taken against the original data.

\subsection{Foundation models}

For experiments on single-cell datasets, the scGPT \cite{cui2023scgpt} whole-human checkpoint is incorporated for message creation in FIMP-scGPT, consisting of a 12-layer transformer with 54 million parameters. scGPT is pretrained using a masked gene expression prediction objective on over 33 million cells from a diverse array of human tissues and organs. The pretrained gene embedding table is also utilized from the pretrained scGPT checkpoint, representing pretrained knowledge about gene identities in transcriptomics datasets. Additionally, we also utilize the gene embeddings obtained by GenePT \cite{chen2023genept}, which are GPT-3.5 embeddings of gene function descriptions based on biomedical literature, as another pretrained gene embedding experiment. For image classification, a standard ViT \cite{dosovitskiy2020image} with 12 transformer layers and 86 million parameters is used as a message creator. The patch encoder from the ViT is also reused from the ViT embedding module. For experiments on fMRI brain recordings, the BrainLM \cite{ortega2023brainlm} model was used, which consists of a Masked Autoencoder transformer with an 8-layer encoder and 4-layer decoder, totaling 26 million parameters.

\subsection{Baselines}

For both supervised and self-supervised tasks, we compare FIMP against popular message-passing GNN architectures, including GCN \cite{kipf2016semi}, GraphSAGE \cite{hamilton2017inductive}, Graph Attention Networks (GATs) \cite{velivckovic2017graph}, and Graph Isomorphism Networks (GINs) \cite{xu2018powerful}. We also compare FIMP against more recent GNN architectures, namely GraphMAE \cite{hou2022graphmae}, a masked graph autoencoder model, and GPS Graph Transformer \cite{rampavsek2022recipe}, a SOTA graph transformer framework. For supervised classification tasks, we additionally compare to the pretrained foundation model with no graph structure input.

\section{Related works}
\label{sec:related_works}

\subsection{Attention-based GNNs and Graph Transformers}

GATs \cite{velivckovic2017graph} first introduced the idea of attention-based GNN architectures, learning attention coefficients between neighboring nodes and performing message-passing with a weighted aggregation of neighboring node embeddings. Graph transformers sought to bring the performance and expressivity of the full transformer architecture into the graph domain by modeling graphs as a sequence of node embeddings that represented a fully-connected graph. Graph Transformer Networks (GTNs) \cite{yun2019graph} proposed the first graph transformer architecture, which could learn new graph structures and multi-hop connections. Graph-BERT \cite{zhang2020graph} proposed pretraining on subgraphs and finetuning for node classification and graph clustering tasks. Graph Transformer \cite{dwivedi2020generalization} proposed utilizing laplacian eigenvectors as positional encodings for node tokens. SAN \cite{kreuzer2021rethinking} improved upon it by introducing learnable spectral positional encodings, and Graphormer \cite{ying2021transformers} further proposed spatial and centrality encodings for nodes to capture structural relation and node importance in graphs. GPS Graph Transformer \cite{rampavsek2022recipe} proposed a general framework for building expressive graph transformers composed of positional and structural encodings, graph features, and GNN and attention layers.

In contrast to these works, FIMP views a single node as sequence of feature tokens similar to transformer input sequences, rather than a single node embedding vector as in GATs and graph transformers. Cross-attention is computed between the node feature sequences of two neighboring nodes, resulting in a message which is passed between the two nodes. FIMP is the first method that uses feature-level attention between tokenized nodes to create a message for message-passing over graphs. The node tokenization procedure allows FIMP to leverage pretrained non-textual foundation models as the message creator, something which other attention-based GNNs and graph transformers are unable to do.




\subsection{LLMs on Text-Attributed Graphs}


More recent works have explored using Large Language Models (LLMs) in conjunction with LLMs on text-attributed graphs. GPT4Graph \cite{guo2023gpt4graph} evaluated LLM reasoning capabilities on graph reasoning tasks, establishing a benchmark of graph-related tasks for language models. Talk Like a Graph \cite{fatemi2023talk} and NLGraph \cite{wang2023can} conducted similar studies exploring graph reasoning capabilities of LLMs, and released the GraphQA and NLGraph benchmark datasets, respectively. One-for-all \cite{liu2023one} used LLMs as an encoding module for text-attributed graphs, and trained a unified GNN model to do node, edge, and graph-level classification using node-of-interest (NOI) subgraphs and prompt nodes. In contrast to these works, we focus on non-textual foundation models and graphs, which have not been explored extensively in graph-based tasks. Our work can be seen as a parallel work to LLM-based works on graphs, aiming to effectively leverage foundation models pretrained on other data domains besides natural language.

\section{Attention Visualizations}
\label{appendix:attention_visuals}

\begin{figure}[!ht]
  \centering
  \includegraphics[width=0.85\linewidth]{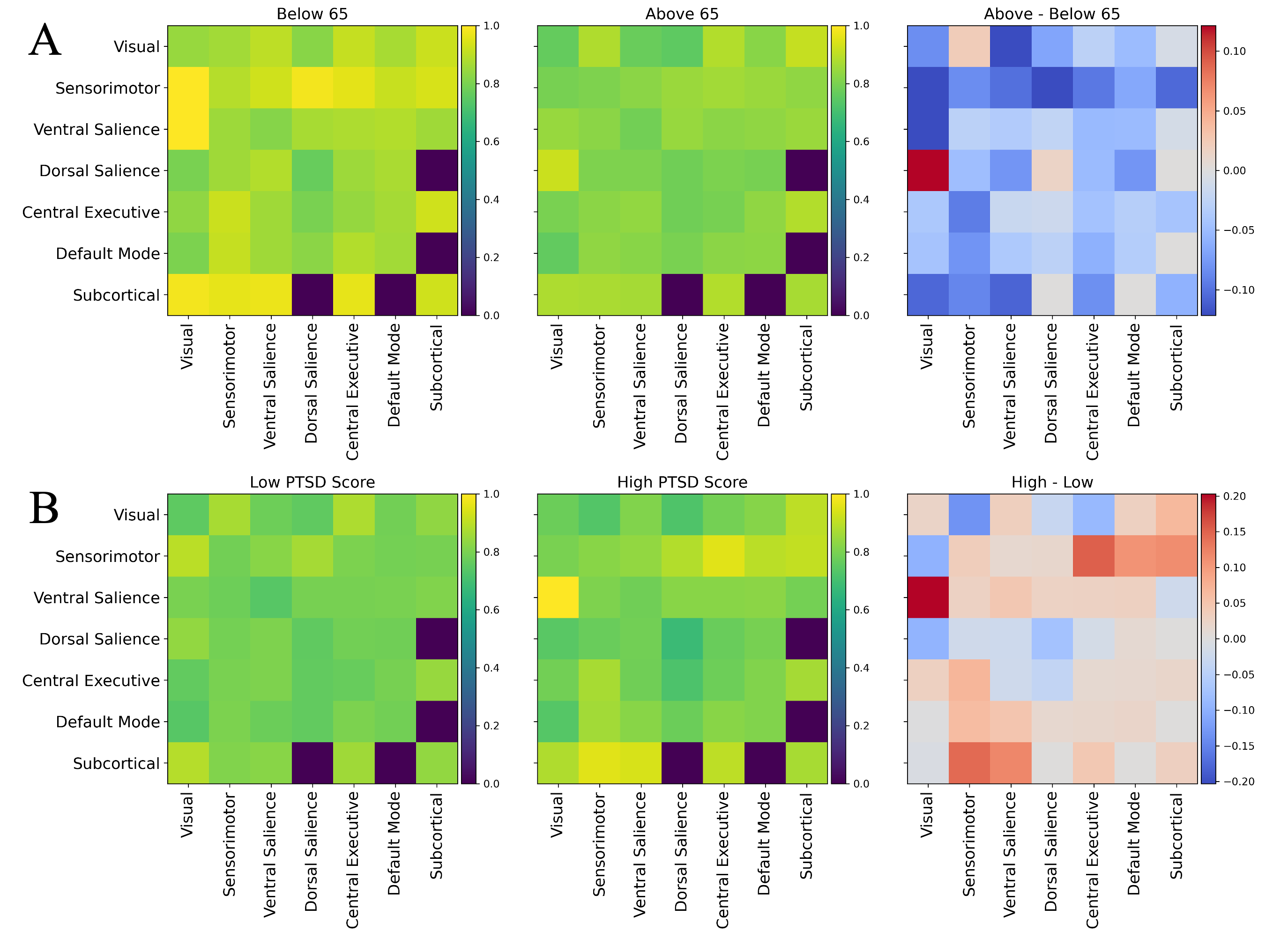}
  \caption{Visualizations of FIMP feature-level attention between different functional groups in the brain. (A) Averaged attention heatmaps between functional regions of the brain for different age populations, with the difference in attention by age group visualized on the right subplot. (B) Similar heatmaps visualized for post-traumatic stress disorder (PTSD) scores, highlighting differences in attention in patients with low vs high PTSD score.}
  \label{fig:ukb_FIMP_attention_panel}
\end{figure}

\subsection{Functional Region Attention in fMRI Recordings}
\label{ssec:supp_fmri_connectivity}

During message passing on the fMRI recording graphs, FIMP generates cross-attention matrices during message-creation between feature tokens of neighboring brain regions in the K-nearest neighbors graph. We group the 424 brain voxels into 7 functional regions, namely the visual, sensorimotor, ventral salience, dorsal salience, central executive, default mode, and subcortical regions of the brain. Taking 100 unseen test set recordings, we extract attention matrices between all connected nodes, average the attention matrices across timepoints per node, and split patient recordings according to conditions such as Age and post-traumatic stress disorder (PTSD) score. We then average attention values across patient recordings with the same condition, and aggregate the node attention into the 7 functional regions, allowing us to examine differences in functional region attention between patients with different conditions.

In Figure \ref{fig:ukb_FIMP_attention_panel}A, the attention between functional regions is shown between patients below 65 years of age (left) and those above 65 (middle). The difference in attention between the two groups, as visualized on the rightmost plot, indicates that older patients tend to have higher attention between the dorsal salience regions and visual cortex regions. This follows previous literature that shows changes in dorsal pathways as people age \cite{yan2023structural}. Furthermore, Figure \ref{fig:ukb_FIMP_attention_panel}B shows similar visualizations for patients with high and low PTSD scores, revealing higher attention between sensorimotor areas and central executive, and subcortical areas. This also follows previous literature on the somatosensory basis of PTSD, where arousal and higher-order capacities get affected \cite{kearney2022brain}. These patterns in attention reveal potential differences in functional region attention picked up by FIMP among patients of varying conditions.

\begin{figure}
  \centering
  \includegraphics[width=0.9\linewidth]{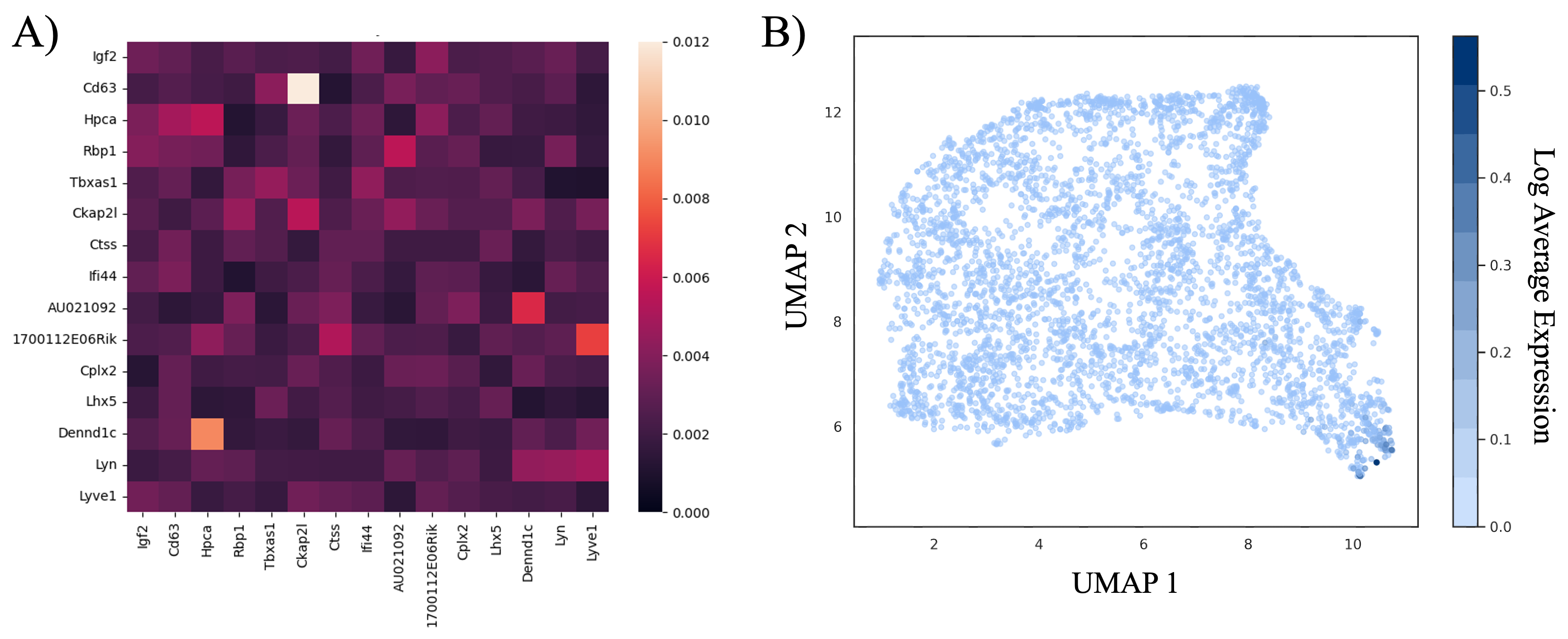}
  \caption{(A) Averaged attention between 15 genes across edges connecting neighboring astrocyte cells in the mouse hippocampus dataset. (B) UMAP of learned gene embeddings from FIMP, colored by average expression value of each gene across astrocyte cells.}
  \label{fig:slideseqv2_figure_panel}
\end{figure}

\subsection{Attention Case Study 2: Gene Interactions in Spatial Transcriptomics}
\label{ssec:supp_spatial_transcriptomics}

In spatial transcriptomics datasets, each node corresponds to a cell which is represented by a set of expressed genes. Message-creation in FIMP provides cross-attention matrices representing interactions between genes of neighboring cells. Gene interactions receiving higher attention between nodes can highlight possible biological connections which can be avenues of potential further exploration in the data. For example, Figure \ref{fig:slideseqv2_figure_panel}A shows an averaged attention heatmap across all self-edges connecting astrocyte cells in a subgraph sampled from the mouse hippocampus dataset \cite{stickels2021highly}. This astrocyte-astrocyte feature-level attention matrix identifies a key interaction between CD63, a member of the tetraspanin family of cell surface proteins, and CKAP2L, a mitotic spindle protein controlling cellular division. Previous work has identified that CD63 may be either pro- or anti-tumorigenic, depending on tissue context \cite{dey2023revisiting}. CD63 expression is also highly enriched in glioblastoma, a highly lethal malignancy of the astrocytes, and may play a role in progression of these cancers \cite{aaberg2018co}. This hints that CD63 may play an important role in controlling cellular division through astrocyte-astrocyte cellular communication, which may represent an exciting new target for antitumoral agents.

Figure \ref{fig:slideseqv2_figure_panel}B shows a UMAP embedding of the gene embeddings learned by FIMP-base during masked gene expression prediction training. Each gene is colored by its average expression value across all astrocyte cells in the mouse hippocampus dataset. We see that the learned embeddings form distinct structures during training, and that highly-expressed genes for astrocytes are clustered together in one region in the bottom-right. We hypothesize that this ability to learn gene vectors in embedding space and contextualize them for different cell types allows FIMP to outperform other methods in gene expression prediction tasks.

\end{document}